\documentclass[10pt,journal,compsoc]{IEEEtran}

\ifCLASSOPTIONcompsoc
  \usepackage[nocompress]{cite}
\else
  \usepackage{cite}
\fi

\usepackage{subcaption}
\usepackage{graphicx}
\usepackage{amsmath}
\usepackage{amssymb}
\usepackage{booktabs}
\usepackage{multirow}
\usepackage{siunitx}
\usepackage{xcolor}

\usepackage{dblfloatfix}

\begin{document}
%
\title{VG-TIE: An interpretable tabular-to-image encoding method based on visibility graphs}

\author{David Chushig-Muzo,
        Luis M. López-Ramos,
        Ángeles Rodríguez de Cara,
        Eva Milara,
        Luis Zhinin-Vera,
        Diego H. Peluffo-Ordóñez
\IEEEcompsocitemizethanks{\IEEEcompsocthanksitem David Chushig-Muzo, Luis Miguel López-Ramos, Ángeles Rodríguez de Cara, E. Milara are with the Department of Signal Theory and Communications, Telematics and Computing Systems, Rey Juan Carlos University (URJC), Madrid, Spain. E-mail: \{david.chushig, luismiguel.lopez, angeles.decara, eva.milara\}@urjc.es
\IEEEcompsocthanksitem Luis Zhinin-Vera is with the LoUISE Research Group, University of Castilla-La Mancha, Albacete, Spain. E-mail: luis.zhinin@uclm.es
\IEEEcompsocthanksitem Diego H Peluffo-Ordoñez is with the School of Mathematical  and Computational Sciences, Yachay Tech University, Ecuador. E-mail: dpeluffo@yachaytech.edu.ec
}
}


\IEEEtitleabstractindextext{%
\begin{abstract}
Tabular-to-image encoding methods enable the application of models based on both convolutional neural networks and vision transformers to tabular data, transforming feature vectors into images. Existing methods employ linear and nonlinear dimensionality reduction techniques (\textit{e.g.,} Principal Component Analysis (PCA), t-SNE, and UMAP) to determine pixel positions, resulting in images whose spatial layout do not inherently reflect feature relationships. This paper introduces Visibility Graphs for Tabular-to-Image Encoding (VG-TIE), a novel method that encodes the structure of feature values using Natural Visibility Graph (NVG) and Horizontal Visibility Graph (HVG) into a two-dimensional space obtained through PCA. The resulting images are model-agnostic and intrinsically interpretable. Each pixel corresponds to an input feature, its intensity reflects the magnitude and direction of deviation from the population mean, and edges represent formally defined visibility relationships between features. VG-TIE provides two interpretability methods: \textit{(i)} feature ranking from node degree distributions; and \textit{(ii)} local and global feature importance from pixel intensity combined with Grad-CAM. Experiments on six public tabular datasets show that VG-TIE is competitive with other tabular-to-image methods while providing interpretability on feature importance and ranking similar to intrinsic interpretable methods. The results highlight the potential of the proposed image-based transformation to provide an effective framework that expands the use of deep learning across tabular data domains.
\end{abstract}

}

\maketitle

\IEEEdisplaynontitleabstractindextext

\IEEEpeerreviewmaketitle

\ifCLASSOPTIONcompsoc
\IEEEraisesectionheading{\section{Introduction}\label{sec:introduction}}
\else
\section{Background and motivation}
\label{sec:introduction}
\fi

\IEEEPARstart{T}{abular} data is one of the most prevalent data formats across domains such as healthcare, finance, and bioinformatics~\cite{borisov2022deep}. Convolutional neural networks (CNNs) and Vision Transformers (ViTs) have achieved remarkable results in image and text domains, but their performance has not been extended to tabular data~\cite{mamdouh2026tab2visual}. CNNs and ViTs are designed to exploit spatial locality and temporal patterns, characteristics that tabular data lacks by definition~\cite{borisov2022deep}. This limitation has motivated a line of work that transforms tabular data into image-like representations to leverage the high predictive performance of CNNs and pretrained ViTs.

Several tabular-to-image encoding methods have been proposed, such as DeepInsight~\cite{sharma2019deepinsight}, TINTO~\cite{castillo2023tinto}, REFINED~\cite{bazgir2020representation}, which use dimensionality reduction methods such as Principal Component Analysis (PCA), t-distributed Stochastic Neighbor Embedding (t-SNE) and the Uniform Manifold Approximation and Projection (UMAP) to assign input features to pixel positions in a two-dimensional image, filling pixel intensities with feature values. These methods generate images in which spatial proximity reflects feature similarity. Consequently, neighboring pixels tend to represent related features in the input space. Tabular-to-image methods have been used in different applications, including disease classification~\cite{lara2025transfer, gomez2026tabular}, indoor localization~\cite{liu2026interpretable, castillo2025mimo, talla2023novel}.

Despite these advances, the resulting spatial arrangement is not inherently interpretable, as identifying the location of a feature within the image requires an external mapping between features and pixels. In critical domains such as healthcare, model interpretability is paramount. Although tabular-to-image encoding methods have obtained promising predictive performance in recent years~\cite{sharma2019deepinsight, zhu2021converting, medeiros2023comparative}, their interpretability remains under study, with a limited of studies beginning to address this issue explicitly~\cite{gomez2024lm, lee2024table2image, mondragon2026interpretable, chushig2026tabsom}.

Visibility graphs, originally proposed to map time series data into graph structures, offer a robust mechanism for capturing data topology and interpreting temporal dynamics~\cite{azizi2024review}. However, their use as an encoding mechanism within tabular-to-image methods has not been systematically investigated. The key insight is to conceptualize the feature values as an ordered representation that inherently captures structural information about feature importance. To the best of our knowledge, no prior work has applied visibility graphs to tabular-to-image encoding, providing model interpretability.

This paper introduces Visibility Graphs for Tabular-to-Image Encoding (VG-TIE), a novel encoding framework that exploits the structural relationships among features to provide interpretable spatial representations. It employs the Horizontal Visibility Graph (HVG)~\cite{luque2009horizontal} and the Natural Visibility Graph (NVG)~\cite{lacasa2008time} to construct the graph, whereas PCA is used to project feature values into a two-dimensional space. VG-TIE provides two interpretability methods: \textit{(i)} feature ranking from node degree distributions; and \textit{(ii)} local and global feature importance obtained by combining pixel intensity with Gradient-weighted Class Activation Mapping (Grad-CAM)~\cite{selvaraju2017grad}. 

The main contributions of this paper are:

\begin{itemize}
    \item A novel tabular-to-image method that encodes feature value structure as a visibility graph with PCA-based approach for node positions in the RGB image.

    \item An effective approach for using tabular samples to create ordered sequences via PCA loading magnitude, enabling the application of visibility graphs such as NVG and HVG.
    
    \item An interpretability framework for visibility-graph-based image representations, combining degree-based feature ranking and both local (sample-level) and global (feature-level) Grad-CAM attribution.

    \item A comprehensive benchmark against tabular-to-image methods, including TINTO, IGTD, DeepInsight, BarGraph, DistanceMatrix, Combination, BIE, Fotomics. 

    \item Experiments on six tabular datasets demonstrate that VG-TIE is competitive with other state-of-the-art methods while providing interpretability, mainly related to feature importance.
    
\end{itemize}

The remainder of the paper is organized as follows. Section~\ref{sec:related_work} presents the related work. Section~\ref{sec:methods} describes VG-TIE and the interpretability methods. Section~\ref{sec:results} presents the main results. Section~\ref{sec:discussion} discusses the findings, limitations and future work, and finally Section~\ref{sec:conclusions} presents the conclusions of the paper.

\section{Related work}
\label{sec:related_work}

Several approaches for transforming tabular data into images have been developed in the literature. For instance, BarGraph represents each sample as a vertical bar chart, with each feature mapped to a fixed column and its bar height determined by the corresponding feature value~\cite{sharma2022classification}. Binary Image Encoding (BIE) and correlated BIE encode numerical values as sequences of binary digits and stack them to produce a two-dimensional image representation~\cite{briner2023tabular}. 

DeepInsight uses t-SNE or kernel PCA to map the data into a two-dimensional space, and then constructs the final image by applying convex hull analysis to determine the image orientation and bounding region~\cite{sharma2019deepinsight}. Several variants have been proposed from its publication. For instance, MRep-DeepInsight generates multiple image representations per sample through manifold techniques to improve predictive performance~\cite{sharma2024enhanced}. DeepInsight-3D converts high-dimensional multi-omics tabular data into tensors~\cite{sharma2023deepinsight}.

Fotomics applies the Fourier transform to each feature, mapping the real and imaginary parts of the resulting complex coefficients to the x- and y-axes of a two-dimensional plane. The resulting image captures the structural characteristics of the features in the Fourier domain~\cite{alenizy2025transforming}. FC-Viz first groups highly correlated features into clusters by analyzing the relationships among representative features from each cluster~\cite{damri2024towards}. Then, it employs ant colony optimization to determine the spatial arrangement of the features. Dimensionality reduction techniques are used to obtain the pixel intensities~\cite{damri2024towards}.

IGTD~\cite{zhu2021converting} and LM-IGTD~\cite{gomez2024lm} perform feature-to-pixel assignment as a distance-preservation optimization problem. Specifically, they obtain two rankings: one based on the pairwise distances among features and the other based on the pairwise distances among samples. Then, they iteratively swap feature assignments to minimize the discrepancy between these two rankings~\cite{zhu2021converting}. LM-IGTD extends IGTD to support low-dimensional and mixed-type data using an unsupervised stochastic noise generation algorithm~\cite{gomez2024lm}. 

REFINED employs t-SNE, kernel PCA, and multidimensional scaling to derive a two-dimensional coordinate for each feature, subsequently mapping feature values to pixel intensities~\cite{bazgir2020representation}. Similarly, TINTO applies PCA and t-SNE to determine the spatial locations of features. The resulting coordinates are then transposed, scaled, and rounded to integer values~\cite{castillo2023tinto}. TINTO additionally incorporates a blurring technique to generate smoother image representations.

\section{Methods}
\label{sec:methods}

\subsection{VG-TIE}

Let $X \in \mathbb{R}^{N \times n}$ be a tabular dataset with $N$ samples and $n$ features, and let $y \in \{0,1\}^{N}$ denote the corresponding vector labels. VG-TIE converts each sample $x_i \in \mathbb{R}^{n}$ into a fixed-size RGB image $I_i \in [0,1]^{H \times W \times 3}$ through a three-stage pipeline: (1) feature ordering by PCA loading magnitude, (2) visibility graph construction from the ordered feature values, (3) pixel layout and image rendering. 

\subsubsection{Feature ordering by principal component analysis loading}

Let $\mathbf{x} \in \mathbb{R}^{n}$ denote a tabular sample with $n$ features. Each feature is normalized using the mean $\mu_j$ and standard deviation $\sigma_j$ of training set:

\begin{equation}
z_j = \frac{x_j - \mu_j}{\sigma_j},
\qquad j = 1, \ldots, n
\end{equation}

This removes scale differences between features so that the subsequent PCA is not dominated by features with large numerical ranges. Missing values are imputed with the column-wise training median prior to normalization. PCA is then applied to the normalized training matrix $\mathbf{Z} \in \mathbb{R}^{N \times n}$. The first eigenvector
\begin{equation}
\mathbf{v}_1 =
(v_{1,1}, \ldots, v_{1,n})^{\top}
\end{equation}
of the sample covariance matrix captures the linear combination of features that explains the largest proportion of variance in the training data. Each component $v_{1,j}$ is the loading of feature $j$ on the first principal component (PC1) and quantifies the contribution of feature $j$ to the dominant pattern of variation observed across training samples.

Features are ranked in descending order according to the magnitude of their PC1 loadings, $|v_{1,j}|$. The rank of feature $j$ is defined as

\begin{equation}
\sigma(j)
=
\operatorname{argsort}\!\left(\left|v_{1,j}\right|\right)_{\downarrow},
\qquad j = 1, \ldots, n
\end{equation}
such that $\sigma(1)$ corresponds to the feature with the largest absolute loading on the PC1 and $\sigma(n)$ to the smallest. The reordered feature vector is

\begin{equation}
\boldsymbol{\psi}
=(\psi_1,\ldots,\psi_n)=\left(z_{\sigma(1)},\ldots,z_{\sigma(n)}\right)
\end{equation}

This ordering places the statistically most informative features at the beginning of the sequence. Note that features with large PC1 loadings occupy early positions in the visibility sequence, where their local maxima exert the greatest influence on long-range edge formation. The PCA ordering is fitted once on the training set and subsequently applied to validation and test samples.

\subsubsection{Visibility graph construction}

Given the ordered feature vector $\boldsymbol{\psi}$, we define a visibility graph $\mathcal{G} = (V,E)$ where the node set $V = \{v_1,\ldots,v_n\}$ contains one node for each feature, and the edge set $E$ is determined by a visibility criterion applied to the ordered sequence. An undirected edge $(v_j,v_k)$ with $j < k$ is included in $E$ if and only if no intermediate node $v_l$ ($j<l<k$) blocks the line of sight between $v_j$ and $v_k$. The blocking depends on the chosen visibility graph variant. We used two approaches: HVG and NVG.

For the HVG, visibility is determined by relative height. An intermediate node $v_l$ blocks the edge $(v_j,v_k)$ whenever

\begin{equation}
\psi_l \geq \min(\psi_j,\psi_k)
\end{equation}
Equivalently, two nodes are connected if and only if
\begin{equation}
\psi_l
<
\min(\psi_j,\psi_k),
\qquad
\forall\, j<l<k
\end{equation}

The HVG has two main characteristics. It is computationally efficient, and the degree of each node depends fully on the local ordinal structure of the sequence, yielding an interpretable graph representation directly linked to the original feature values. Local maxima generally accumulate low degree, whereas local minima tend to accumulate higher degree because they remain visible to a larger number of neighbouring nodes. 

In contrast to the HVG, visibility in the NVG is determined geometrically. An intermediate node $v_l$ blocks the edge $(v_j,v_k)$ whenever the point $(l,\psi_l)$ lies above the straight line segment connecting $(j,\psi_j)$ and $(k,\psi_k)$. Formally, the edge $(v_j,v_k)$ exists if and only if

\begin{equation}
\psi_l
<
\psi_j
+
(\psi_k-\psi_j)
\frac{l-j}{k-j},
\qquad
\forall\, j<l<k
\end{equation}

The NVG is strictly more connected than the HVG because every HVG edge is also an NVG edge, whereas additional long-range edges may be introduced when visibility is preserved along a sloped line. Consequently, the NVG construction has complexity $\mathcal{O}(n^2)$ but it produces denser edge sets than the HVG, encoding a richer set of geometric relationships between feature values.

For both visibility graph variants, each edge $(v_j,v_k)\in E$ is assigned a scalar weight that encodes the joint magnitude of the connected features:

\begin{equation}
w_{jk}
=
\tanh\!\left(
\frac{\left|\psi_j \psi_k\right|}{2}
\right),
\qquad
w_{jk}\in(0,1)
\end{equation}

The term $|\psi_j\psi_k|$ captures the combined magnitude of the two normalized feature values, while the hyperbolic tangent transformation maps the result smoothly into the interval $(0,1)$. As a result, edges connecting two large-magnitude features receive weights close to one and appear as prominent structures in the image representation, whereas edges involving near-zero feature values receive lower weights and contribute less visual intensity.

\subsubsection{Pixel layout and image rendering}

Each feature $j$ is assigned a fixed pixel position $(r_j, c_j)$ derived from its PCA loading vector $V_j$. The loadings are linearly normalized to the image coordinate range $[0, H-1] \times [0, W-1]$. Features mapping to identical pixels are resolved by a deterministic jitter of $\pm 1$ pixel based on feature index. 

The RGB image $I_i$ is rendered in three layers. The blue channel encodes graph connectivity (edge weights) while the green and red channels encode feature magnitude and sign together. Visibility edges are drawn as straight lines between node positions $(r_j, c_j)$ and $(r_k, c_k)$ for each non-zero entry of $A_i$, with blue channel intensity proportional to $|w_{jk}|$.

Feature nodes are rendered as Gaussian-softened circles centered at $(r_j, c_j)$ with radius proportional to $|z_{ij}|$. Positive deviations ($z_{ij} > 0$) activate the green channel; negative deviations activate the red channel. The intensity of both is proportional to $|z_{ij}|$, clamped to $[0,1]$. This RGB encoding ensures that a black pixel means a feature is at its mean, a bright green pixel means a feature is unusually high, and a bright red pixel means a feature is unusually low.

\subsection{Interpretability via visibility graph structure and class activation mapping methods}

A key advantage of VG-TIE over existing tabular-to-image methods is that the visibility graph encoding provides interpretability properties that are directly grounded in the graph structure.

\subsubsection{Feature ranking based on node degree distribution of visibility graphs}

Let $\mathcal{G}_i = (V, E_i)$ denote the horizontal visibility graph constructed for sample $x_i$, where each node $v_j \in V$ corresponds to feature $j$ and edges $E_i$ are determined by the NVG or HVG visibility condition. The degree of node $v_j$ in sample $i$ is defined as follows:
\begin{equation}
d_j^{(i)} = \sum_{k \neq j} \mathbf{1}[A_i[\sigma(j), \sigma(k)] \neq 0]    
\end{equation}

where $A_i$ is the adjacency matrix of $\mathcal{G}_i$ and $\sigma(\cdot)$ denotes the position of a feature in the PCA-based visibility ordering. For each feature $j$, we collect the degree distribution $\{d_j^{(i)}\}_{i=1}^{N}$ across all samples and compute three distributional statistics: the mean connectivity $\mu_j$, variability $\sigma_j$, and asymmetry of the degree distribution $\gamma_j$, which are defined:

\[
\mu_j = \frac{1}{N}\sum_{i=1}^{N} d_j^{(i)}
\]

\[
\sigma_j =
\sqrt{\frac{1}{N}\sum_{i=1}^{N}
\left(d_j^{(i)}-\mu_j\right)^2}
\]

\[
\gamma_j =
\frac{1}{N\sigma_j^{3}}
\sum_{i=1}^{N}
\left(d_j^{(i)}-\mu_j\right)^3
\]

Features that consistently act as local peaks in the visibility ordering tend to block their neighbours, yielding low mean degree, high variance, and negative skewness. We combine these statistics into a single importance score:

\begin{equation}
s_j =
\frac{1}{3}
\left(
\tilde{\mu}_j^{-}
+
\tilde{\sigma}_j
+
\tilde{\gamma}_j^{-}
\right)
\end{equation}

where $\tilde{\cdot}$ denotes min--max normalization to the interval $[0,1]$, and $\tilde{\mu}_j^{-} = 1-\tilde{\mu}_j$, $\tilde{\gamma}_j^{-} = 1-\tilde{\gamma}_j$. Note that low mean degree and negative skewness are indicative of structural importance. To obtain ranking, features are ranked in descending order of $s_j$.

\subsubsection{Local and global feature importance via Grad-CAM and pixel intensity}

Given a trained CNN-based model $f_{\theta}$, we apply Grad-CAM to obtain a spatial relevance heatmap $\mathcal{H}^{(i)} \in \mathbb{R}^{H \times W}$ for sample $x_i$. Let $\mathbf{A}^{l} \in \mathbb{R}^{C \times H_l \times W_l}$ denote the activation maps of the last convolutional layer $l$, and let $y^{c}$ denote the score for class $c$. The Grad-CAM weights are defined as

\begin{equation}
\alpha_k^{c}
=
\frac{1}{H_l W_l}
\sum_{u,v}
\frac{\partial y^{c}}
{\partial A_{k,u,v}^{l}}    
\end{equation}

and the heatmap is computed as

\begin{equation}
\mathcal{H}^{(i)}
=
\mathrm{ReLU}
\!\left(
\sum_{k}
\alpha_k^{c}
\, \mathbf{A}_k^{l}
\right)    
\end{equation}

followed by normalization to the interval $[0,1]$. Since VG-TIE places each feature $j$ at a fixed pixel position $(r_j, c_j)$ determined by the PCA, we can directly attribute the heatmap intensity back to the original features. We additionally weight by the pixel intensity $\mathcal{I}^{(i)}_{r_j,c_j}$ of the image, which encodes the magnitude of the feature's standardized deviation:

\begin{equation}
\phi_j^{(i)}
=
\mathcal{H}^{(i)}_{r_j,c_j}
\times
\mathcal{I}^{(i)}_{r_j,c_j}    
\end{equation}

where $\mathcal{I}^{(i)}=\frac{1}{3}
\sum_{c=1}^{3}I_c^{(i)}$ is the mean intensity across RGB channels. This product fuses spatial relevance (Grad-CAM) with pixel intensity, yielding a local feature attribution score $\phi_j^{(i)}$ for each feature in each sample.

The local attribution scores $\phi_j^{(i)}$ are aggregated across the entire set of images to produce a global feature importance ranking grounded in what the CNN actually attended to across all samples. We define three global quantities.

The global importance $\Phi_j$ is the mean attribution across all $N$ samples:
\begin{equation}
\Phi_j = \frac{1}{N} \sum_{i=1}^{N} \phi_j^{(i)}    
\end{equation}

The per-class importance $\Phi_j^{(c)}$ averages only samples belonging to class $c$:

\begin{equation}
\Phi_j^{(c)} =
\frac{1}{N_c}
\sum_{i:\, y_i = c}
\phi_j^{(i)},
\qquad
N_c = \left|\{\, i : y_i = c \,\}\right|    
\end{equation}

The discriminative score $\Delta_j$ measures which features the CNN attends to differently across classes:

\begin{equation}
\Delta_j = \Phi_j^{(1)} - \Phi_j^{(0)}
\end{equation}

A large positive $\Delta_j$ indicates that feature $j$ attracts more attention in class~1 samples; a large negative value indicates preference for class~0. Features with $|\Delta_j|$ close to zero are attended to equally regardless of class and are therefore less discriminative.

\section{Results}
\label{sec:results}

In this section, we present the datasets used, experimental setup as well as the classification results compared VG-TIE with other tabular-to-image methods. Finally, we show the feature maps, and the local and global interpretability using our method.

\subsection{Datasets}
\label{sec:dataset_description}

The evaluated datasets include Musk 1 (MUSK), Parkinson's Disease (PAR), Pima Indians Diabetes (PIMA), QSAR Biodegradation (QSAR), Spambase (SPA) and Wisconsin Diagnostic Breast Cancer (WDBC). These datasets are from the UCI Machine Learning Repository~\footnote{UCI Repository: http://archive.ics.uci.edu/ml/}. Table~\ref{table:dataset_summary} summarizes the datasets, including the total number of samples and features. 

\begin{table}[htbp]
\centering
\caption{Datasets used in this study.}
\label{table:dataset_summary}
\scalebox{0.95}{
\begin{tabular}{llrr}
\toprule
Identifier   & Dataset                          & Samples  & Features  \\ \midrule
MUSK & Musk version 1                   & 476  & 168                \\
PAR  & Oxford Parkinson's Disease       & 195  & 22                \\
PIMA & Pima Indians Diabetes            & 768  & 8                 \\
QSAR  & QSAR Biodegradation              & 1055 & 41                \\
SPA  & Spambase                         & 4601 & 57                \\
WDBC  & Wisconsin Breast Cancer          & 569  & 30                \\
\bottomrule
\end{tabular}
}

\end{table}

\subsection{Experimental setup}
\label{sec:experimental_setup}

We split each dataset into two independent subsets, a training subset (80\% samples) and test subset (20\% samples). To evaluate the generalization capability of predictive models, all methods are evaluated under 5-fold stratified cross-validation. Class imbalance was addressed through random undersampling, and it is applied only for training subset to prevent data leakage. For assessing classification performance, we considered sensitivity, specificity, and the Area Under the Receiver Operating Characteristic (AUCROC), averaged over five folds.

We compare VG-TIE against different tabular-to-image methods, including TINTO (t-SNE, PCA, and blur variants), IGTD, DeepInsight (t-SNE and PCA variants), BarGraph, DistanceMatrix, Combination, BIE, and Fotomics. Images are generated at $64\times64$ to ensure identical input dimensionality to the CNN. A lightweight CNN, consisting of three convolutional blocks followed by two fully connected layers with dropout, is trained for up to 60 epochs with early stopping (patience=10) using the Adam optimizer. All methods use the same SmallCNN backbone with default hyperparameters to ensure a fair comparison that isolates image generation quality as the only factor.

\subsection{Classification results}

\begin{table*}[!t]
\centering
\caption{Benchmark of tabular-to-image methods using the mean AUCROC accompanied with the corresponding standard deviation. The \textbf{bold} values indicate the highest AUCROC for each dataset, whereas the \textit{italic} values indicate the second-highest AUCROC values.}
\label{table:benchmark_t2i_methods}
\begin{tabular}{lcccccc}
\toprule
Method & PIMA & PAR & WDBC & QSAR & SPA & MUSK \\ \midrule

BarGraph &
$0.8114 \pm 0.0328$ &
$\mathit{0.9069 \pm 0.0336}$ &
$0.9911 \pm 0.0035$ &
$0.8912 \pm 0.0204$ &
$0.9707 \pm 0.0032$ &
$0.8748 \pm 0.0665$ \\

BIE &
$0.7283 \pm 0.0828$ &
$0.7883 \pm 0.0506$ &
$0.9792 \pm 0.0055$ &
$0.8611 \pm 0.0228$ &
$\mathit{0.9766 \pm 0.0030}$ &
$0.6041 \pm 0.0505$ \\

Combination &
$\mathbf{0.8183 \pm 0.0265}$ &
$0.9164 \pm 0.0423$ &
$0.9886 \pm 0.0079$ &
$0.9140 \pm 0.0182$ &
$\mathbf{0.9767 \pm 0.0026}$ &
$0.8923 \pm 0.0642$ \\

DeepInsight (tSNE) &
$0.6276 \pm 0.0643$ &
$0.8322 \pm 0.0911$ &
$0.5828 \pm 0.1960$ &
$0.4583 \pm 0.1544$ &
$0.4535 \pm 0.2199$ &
$0.7070 \pm 0.0799$ \\

DeepInsight (UMAP) &
$0.5538 \pm 0.0631$ &
$0.6158 \pm 0.2597$ &
$0.5304 \pm 0.2978$ &
$0.8165 \pm 0.0852$ &
$0.3481 \pm 0.1227$ &
$0.7773 \pm 0.0322$ \\

DistanceMatrix &
$0.7087 \pm 0.0607$ &
$0.8224 \pm 0.1504$ &
$0.9484 \pm 0.0184$ &
$\mathbf{0.9165 \pm 0.0150}$ &
$0.9737 \pm 0.0025$ &
$0.8791 \pm 0.0536$ \\

Fotomics &
$0.5440 \pm 0.0634$ &
$0.6559 \pm 0.2264$ &
$0.8157 \pm 0.2213$ &
$0.7957 \pm 0.0795$ &
$0.7388 \pm 0.2011$ &
$0.5396 \pm 0.0733$ \\

IGTD &
$0.5638 \pm 0.0436$ &
$0.5404 \pm 0.2510$ &
$0.7679 \pm 0.1501$ &
$0.6369 \pm 0.0733$ &
$0.9310 \pm 0.0094$ &
$0.4211 \pm 0.0918$ \\

TINTO (PCA) &
$0.5112 \pm 0.1057$ &
$0.6165 \pm 0.3730$ &
$0.5676 \pm 0.3332$ &
$0.8062 \pm 0.0434$ &
$0.7232 \pm 0.1896$ &
$0.7631 \pm 0.0356$ \\

TINTO (tSNE) &
$0.5863 \pm 0.0861$ &
$0.6228 \pm 0.1992$ &
$0.5733 \pm 0.1373$ &
$0.7346 \pm 0.0804$ &
$0.7914 \pm 0.1444$ &
$0.7342 \pm 0.0540$ \\

TINTO (tSNE and blur) &
$0.5645 \pm 0.0936$ &
$0.5138 \pm 0.2352$ &
$0.4916 \pm 0.1945$ &
$0.7290 \pm 0.0350$ &
$0.7951 \pm 0.1484$ &
$0.6495 \pm 0.0738$ \\

\textbf{VG-TIE (HVG)} &
$\mathit{0.8138 \pm 0.0335}$ &
$\mathbf{0.9277 \pm 0.0453}$ &
$\mathbf{0.9945 \pm 0.0034}$ &
$\mathit{0.9145 \pm 0.0109}$ &
$0.9759 \pm 0.0057$ &
$\mathbf{0.9401 \pm 0.0280}$ \\

\textbf{VG-TIE (NVG)} &
$0.8017 \pm 0.0339$ &
$0.9065 \pm 0.0739$ &
$\mathit{0.9943 \pm 0.0048}$ &
$0.8994 \pm 0.0075$ &
$0.9756 \pm 0.0042$ &
$\mathit{0.9283 \pm 0.0214}$ \\

\bottomrule
\end{tabular}

\footnotesize{Description of acronyms: Pima Indians Diabetes (PIMA), Parkinson's Disease (PAR), Wisconsin Diagnostic Breast Cancer (WDBC), QSAR Biodegradation (QSAR), Spambase (SPA), Musk 1 (MUSK).}

\end{table*}

Table~\ref{table:benchmark_t2i_methods} shows the classification results using six datasets and different tabular-to-image methods. VG-TIE (HVG) and VG-TIE (NVG) achieved competitive performance across all datasets. For PIMA, VG-TIE (HVG) obtained the second-highest AUCROC, demonstrating that the visibility graph image representation captures discriminative patterns from tabular data. For WDBC, VG-TIE (HVG) and VG-TIE (NVG) present AUC results over 0.9. For PAR, VG-TIE (HVG) and Combination present best AUC results. For the QSAR and SPA datasets, VG-TIE (HVG) was second and third best-performing method. The NVG and HVG approaches present similar classification results.

\subsection{Feature maps and fingerprints}

\begin{figure*}[h]
    \centering
    \includegraphics[width=0.82\textwidth]{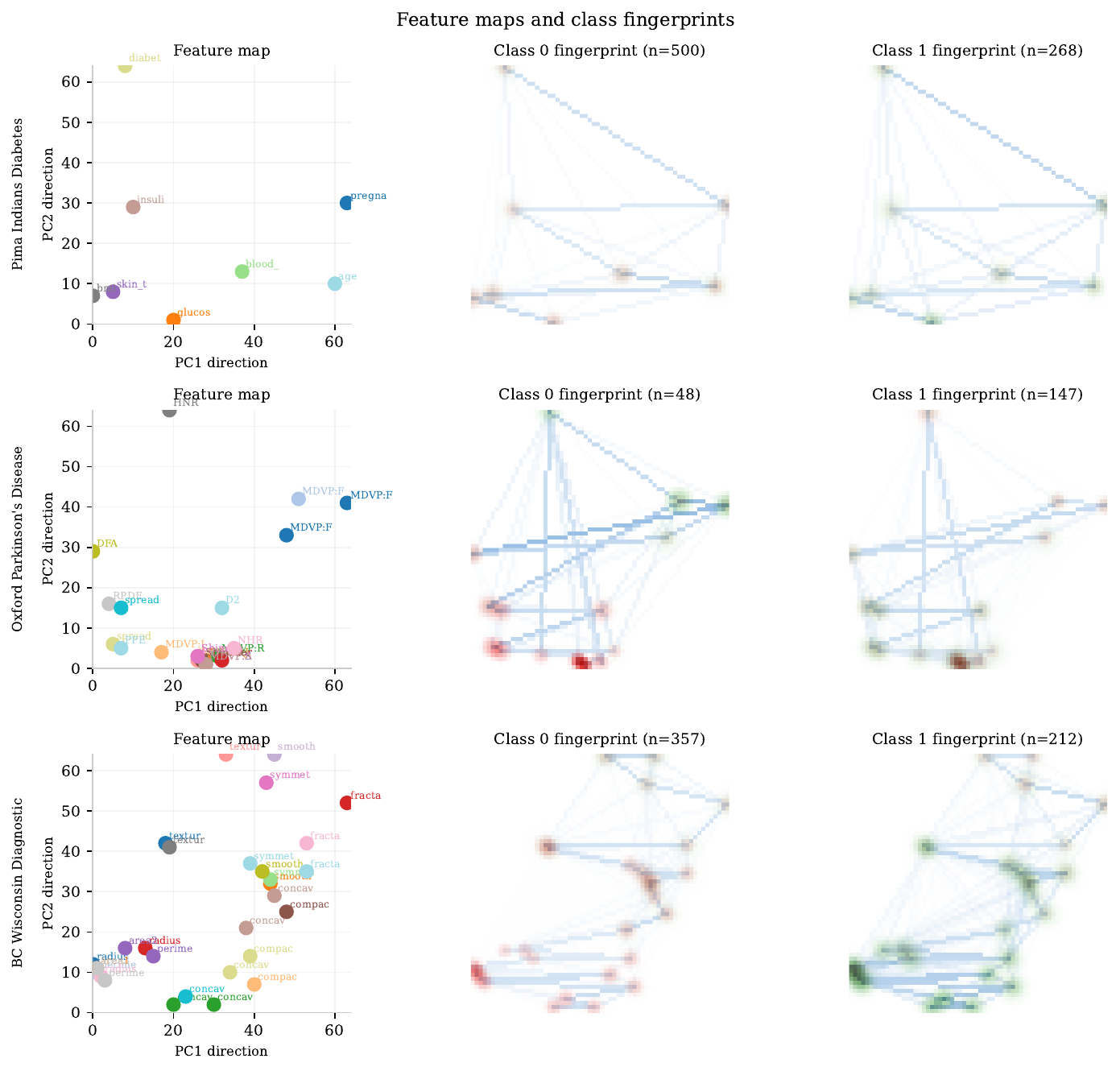} 
    \caption{Feature maps and class fingerprints for three datasets. Left panels: the PCA-derived pixel layout showing where each feature is placed in the image. Middle and right panels: class-averaged VG-TIE (HVG) images obtained by averaging all training samples in each class. Blue lines represent mean edge intensity; green and red blobs represent mean positive and negative feature values respectively.}
    \label{fig:feature_maps}
\end{figure*}

Figure~\ref{fig:feature_maps} shows the feature map and class fingerprints for three representative datasets. In the feature map (left panels), each dot represents one feature placed at its PCA-derived pixel coordinate. The class fingerprints (center and right panels) reveal discriminative structure that differs between classes. 

For PIMA, the class 1 fingerprint (diabetic) shows denser edge patterns in the top-right region of the image, indicating that diabetic samples tend to have higher glucose and age values, producing more long-range edges from those prominent nodes. For PAR, the class 1 fingerprint shows stronger red node blobs in the lower portion, corresponding to negative values of several MDVP features that are lower in Parkinson's patients, alongside denser connectivity in the center of the image. The class 0 fingerprint is sparser overall, consistent with the lower feature variability in the healthy group. For WDBC, the class 1 fingerprint shows green blobs in the lower-left region, corresponding to the size features (radius, area, perimeter) that are consistently elevated in malignant samples. 

\subsection{Local interpretability using Grad-CAM}

\begin{figure*}
  \centering
  \begin{subfigure}[b]{0.9\linewidth}
    \includegraphics[width=\linewidth]{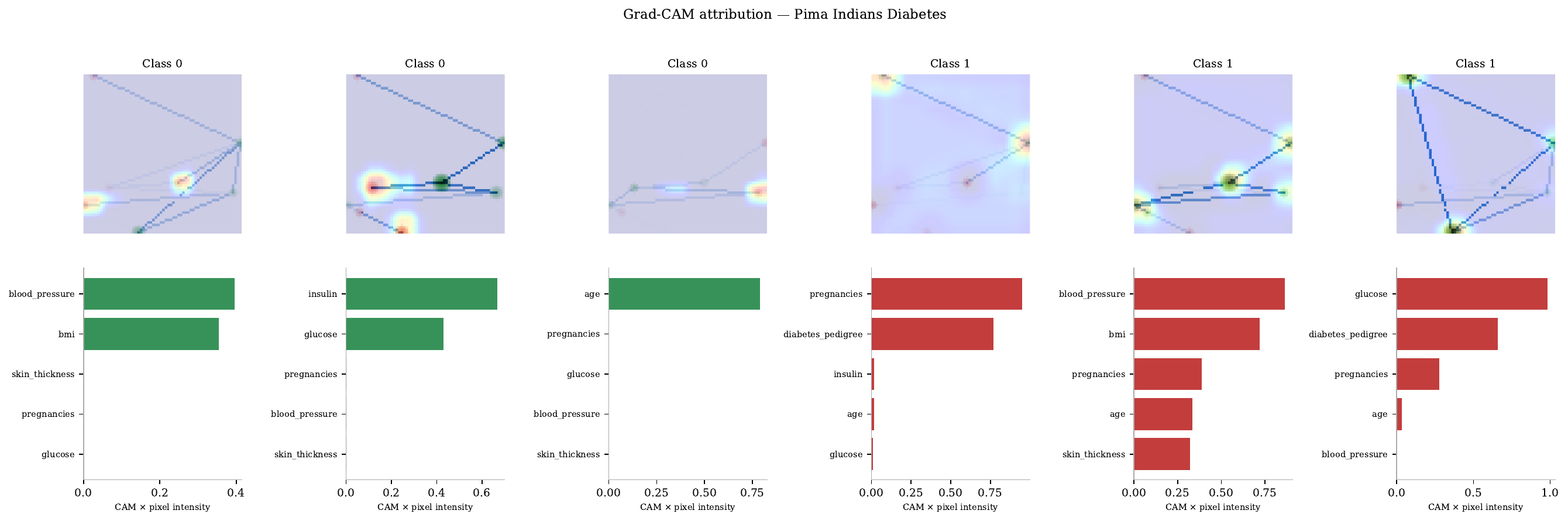} 
    \caption{} 
    \label{fig:gradcam_local:a} 
  \end{subfigure}
  \begin{subfigure}[b]{0.9\linewidth}
    \includegraphics[width=\linewidth]{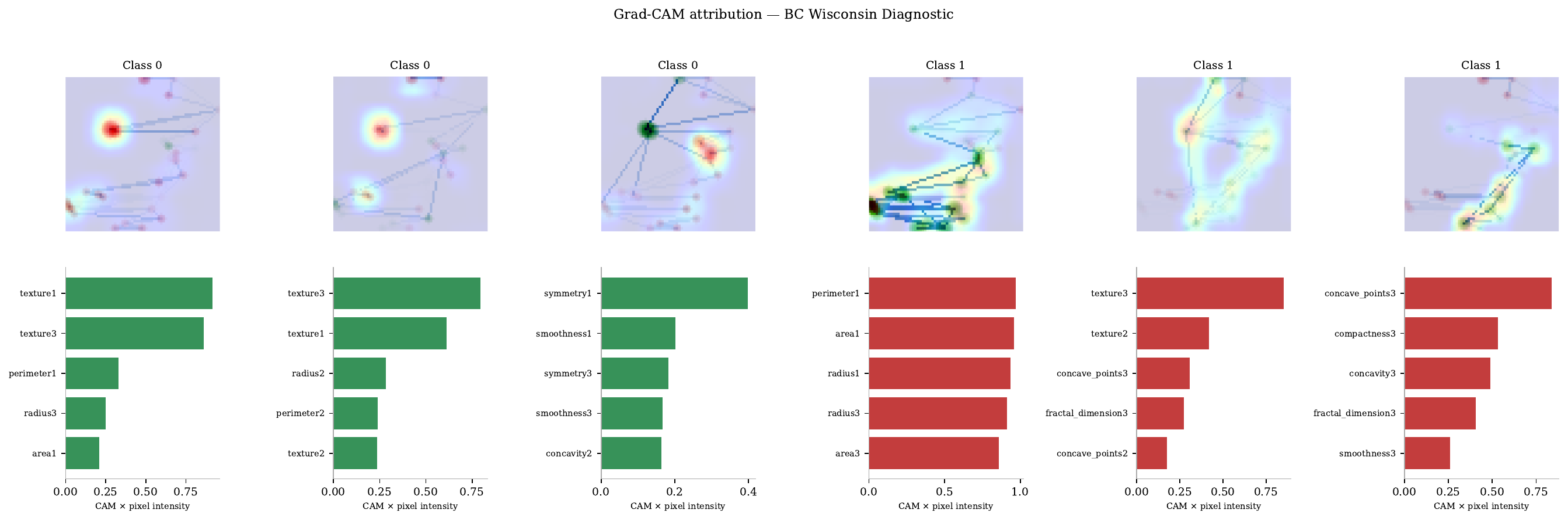} 
    \caption{} 
    \label{fig:gradcam_local:b} 
  \end{subfigure}
  \captionsetup{justification=justified, singlelinecheck=false, margin=6pt} 
      \caption{Local interpretability maps via Grad-CAM for the PIMA and WDBC. Top panels show VG-TIE (HVG) image with Grad-CAM heatmap overlaid whereas bottom panels present top-5 ranked features according to global Grad-CAM importance.}
  \label{fig:gradcam_local} 
\end{figure*}

Grad-CAM computes a class-discriminative heatmap by weighting the activation maps of the last convolutional layer by the global average of their gradients with respect to the target class score. The resulting heatmap highlights spatial regions of the input image that most influenced predictions. Figure~\ref{fig:gradcam_local} presents local Grad-CAM attributions for samples from PIMA and WDBC, showing that VG-TIE provides sample-level explanations directly grounded in the visibility graph structure. 

Figure~\ref{fig:gradcam_local}~(a) presents local Grad-CAM attributions for six individual samples from PIMA. For class 0 (non-diabetic), attribution patterns vary across the three benign samples, reflecting that the CNN uses different feature combinations to confirm non-diabetic status. For class 1 (diabetic), diabetic samples show stronger, more spatially extended heatmaps. 
The features pregnancies and diabetes\_pedigree dominate the fourth sample; blood\_pressure and bmi the fifth, with notably higher attribution values than in the corresponding non-diabetic sample. The sixth sample is the most striking: glucose dominates with attribution approaching 1.0 and bright edges radiating from its pixel position, which is consistent with fasting glucose being the primary clinical criterion for diabetes diagnosis.

Figure~\ref{fig:gradcam_local}~(b) presents local Grad-CAM attributions for six individual samples from WDBC. The samples belonging to class 0 (benign) show a consistent pattern: the CNN focuses mainly to texture features. In the first two benign samples, texture1 and texture3 dominate the attribution ranking. The third benign sample shifts attention toward symmetry1, smoothness1, and symmetry3, with the heatmap showing two activation peaks corresponding to the symmetry and smoothness pixel positions. The malignant samples show a different pattern dominated by size features (perimeter1, area1, radius1, area3, radius3) consistent with the known association between tumour size and malignancy. Heatmaps spread across the lower-left region where size measurements are placed, with bright visibility edges connecting co-elevated size nodes. 

\subsection{Global interpretability using visibility graph and Grad-CAM}

\begin{figure*}
  \centering
  \begin{subfigure}[b]{0.8\linewidth}
    \includegraphics[width=\linewidth]{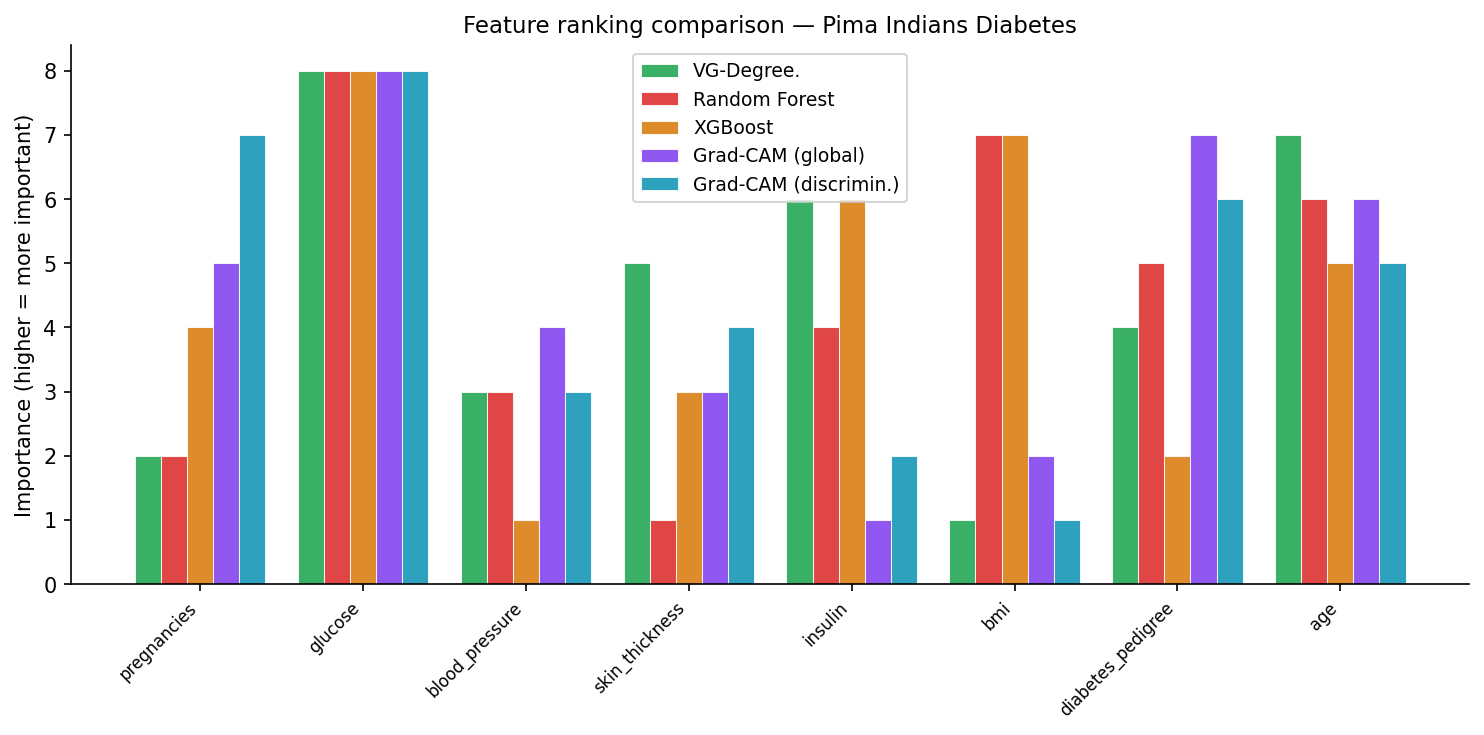} 
    \caption{} 
    \label{fig:gradcam_local:a} 
  \end{subfigure}
  \begin{subfigure}[b]{0.9\linewidth}
    \includegraphics[width=\linewidth]{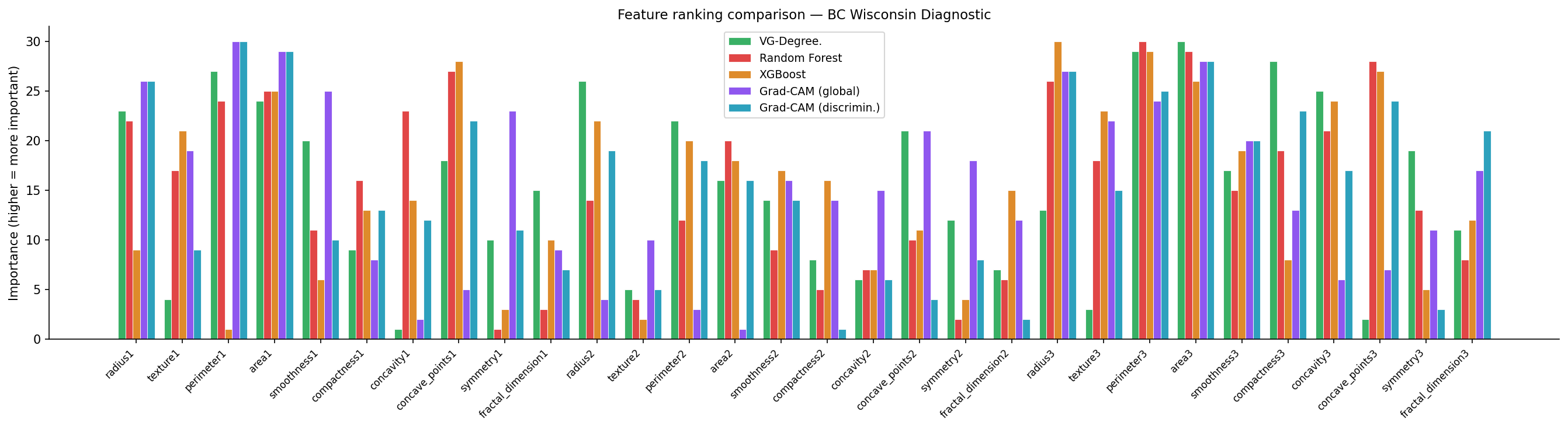} 
    \caption{} 
    \label{fig:gradcam_local:b} 
  \end{subfigure}
  \captionsetup{justification=justified, singlelinecheck=false, margin=6pt} 
      \caption{Feature importance rankings for: (a) PIMA and (b) WDBC. Comparison of five methods: Visibility graph-degree distribution, Random Forest, XGBoost, Grad-CAM global (global), and Grad-CAM (discriminative).}
  \label{fig:global_interpretability} 
\end{figure*}

Figure~\ref{fig:global_interpretability} shows the feature importance rankings produced by five methods for PIMA and WDBC, where each bar represents the rank of a feature. For PIMA (see Figure~\ref{fig:global_interpretability}~(a)), the five methods show consensus on glucose, which is ranked first by all methods simultaneously. This is meaningful since fasting glucose is the primary diagnostic criterion for diabetes. Beyond glucose, methods diverge. The degree distribution assigns insulin and skin thickness relatively high structural ranks, whereas Random Forest and XGBoost rank them lower, suggesting these features create prominent visibility graph patterns without being strongly discriminative for tree-based classifiers. Pregnancies shows the reverse pattern: low degree and tree-based ranks but high Grad-CAM discriminative rank, indicating the CNN attends to it differentially between classes despite its weak structural presence in the graph.

For PIMA (see Figure~\ref{fig:global_interpretability}~(b)), all five methods consistently rank worst-case measurements (suffix\_3) above mean measurements (suffix\_1) and standard error measurements (suffix\_2), with perimeter3, area3, and radius3 reaching maximum ranks across all methods. This cross-method consensus on size features is the strongest agreement observed across both datasets. The main divergence concerns texture features: texture3 ranks highly under both Grad-CAM variants but considerably lower under degree distribution, indicating the CNN exploits texture variation in the image beyond what the HVG topology encodes. Fractal dimension features rank consistently last across all methods regardless of the importance paradigm, confirming they carry negligible discriminative information for this dataset. The degree distribution shows the broadest spread across feature groups, suggesting the HVG structural ranking captures a somewhat different signal from both tree-based and gradient-based methods,  complementary rather than redundant.

\section{Discussion}
\label{sec:discussion}

In this paper, we proposed the interpretable tabular-to-image method named VG-TIE and compared with other state-of-the-art methods. VG-TIE achieves competitive classification performance across all evaluated datasets. The comparison with tabular-to-image methods reveals two insighs. Methods based on deterministic and position-preserving encodings (BarGraph, Combination, BIE, DistanceMatrix) perform similar to VG-TIE because they produce spatially consistent images. Methods based on stochastic dimensionality reduction, such as TINTO, DeepInsight, IGTD, and Fotomics, perform worse on smaller datasets and with few number of features. VG-TIE also demonstrated well performance on datasets with complex feature interactions that benefit from spatial encoding, where the visibility graph captures co-occurrence structure between features. 

The interpretability analysis demonstrates that VG-TIE provides informative local and global explanations, leveraging the graph structure and Grad-CAM. The feature ranking based on degree-based distribution identified features that consistently act as structural peaks in the visibility graph, features whose high or low values create distinctive edge patterns across training samples. For instance, in WDBC, perimeter, area, and radius features rank highest by degree distribution, consistent with the clinical finding that tumour size is the primary indicator of malignancy. The feature ranking comparison across six methods (degree distribution, Random Forest, XGBoost, Grad-CAM and Grad-CAM) showed that the graph-based rankings (degree distribution) agree moderately with model-based rankings, with the strongest agreement occurring for the top-ranked features and the largest disagreement for mid-ranked features. This suggests that the visibility graph structure captures patterns complementary to both tree-based feature importance values and CNN gradient attributions, providing additional interpretability.

\section{Conclusions}
\label{sec:conclusions}

This paper presented VG-TIE, a tabular-to-image method that encodes tabular data as RGB images using visibility graphs (including HVG and NVG) built on PCA. Both the HVG and NVG variants were among the best performing methods, with HVG producing sparser, more interpretable graphs and generally achieving slightly higher AUCROC. VG-TIE extends beyond the tabular encoding by providing predictive modeling and explainability within a single framework. VG-TIE provides interpretability methods based on degree-based structural rankings, global Grad-CAM importance, and local sample-level attributions. The degree-based ranking and global Grad-CAM ranking show moderate agreement, with divergences identifying features whose structural prominence in the graph differs from their discriminative weight in the CNN.


%



\ifCLASSOPTIONcompsoc
  \section*{Acknowledgments}
\else
  \section*{Acknowledgment}
\fi

The authors would like to thank Francisco J. Lara Abelenda for insightful comments and technical feedback.

\ifCLASSOPTIONcaptionsoff
  \newpage
\fi



%

\bibliographystyle{vancouver}
\bibliography{bare_adv_literature}

@article{lacasa2008time,
  title={From time series to complex networks: The visibility graph},
  author={Lacasa, Lucas and Luque, Bartolo and Ballesteros, Fernando and Luque, Jordi and Nuno, Juan Carlos},
  journal={Proceedings of the National Academy of Sciences},
  volume={105},
  number={13},
  pages={4972--4975},
  year={2008},
  publisher={National Academy of Sciences}
}

@article{luque2009horizontal,
  title={Horizontal visibility graphs: Exact results for random time series},
  author={Luque, Bartolo and Lacasa, Lucas and Ballesteros, Fernando and Luque, Jordi},
  journal={Physical Review E—Statistical, Nonlinear, and Soft Matter Physics},
  volume={80},
  number={4},
  pages={046103},
  year={2009},
  publisher={APS}
}

@article{chushig2026tabsom,
  title={TabSOM: A tabular-to-image encoding method based on self-organizing maps},
  author={Chushig-Muzo, David and de Cara, Mar{\'\i}a {\'A}ngeles Rodr{\'\i}guez and Milara, Eva and Lara-Abelenda, Francisco J and Zhinin-Vera, Luis and Peluffo-Ord{\'o}{\~n}ez, Diego H},
  journal={arXiv preprint arXiv:2608.13513},
  year={2026}
}

@article{gomez2024lm,
  title={LM-IGTD: a 2D image generator for low-dimensional and mixed-type tabular data to leverage the potential of convolutional neural networks},
  author={G{\'o}mez-Mart{\'\i}nez, Vanesa and Lara-Abelenda, Francisco J and Peiro-Corbacho, Pablo and Chushig-Muzo, David and Granja, Conceicao and Soguero-Ruiz, Cristina},
  journal={arXiv preprint arXiv:2406.14566},
  year={2024}
}

@article{mondragon2026interpretable,
  title={Interpretable CNN--KAN hybrid architectures for tabular data with synthetic image encoding},
  author={Mondragon-Ruiz, Giovanny and Liu, Jiayun and Castillo-Cara, Manuel and Garc{\'\i}a-Castro, Ra{\'u}l},
  journal={Information Processing \& Management},
  volume={63},
  number={8},
  pages={104954},
  year={2026},
  publisher={Elsevier}
}

@article{gomez2026tabular,
  title={Tabular-to-Image Encoding Methods for Melanoma Detection: A Proof-of-Concept},
  author={G{\'o}mez-Mart{\'\i}nez, Vanesa and Chushig-Muzo, David and Soguero-Ruiz, Cristina},
  journal={Applied Sciences},
  volume={16},
  number={5},
  pages={2459},
  year={2026},
  publisher={MDPI}
}

@article{zhu2021converting,
  title={Converting tabular data into images for deep learning with convolutional neural networks},
  author={Zhu, Yitan and Brettin, Thomas and Xia, Fangfang and Partin, Alexander and Shukla, Maulik and Yoo, Hyunseung and Evrard, Yvonne A and Doroshow, James H and Stevens, Rick L},
  journal={Scientific reports},
  volume={11},
  number={1},
  pages={11325},
  year={2021},
  publisher={Nature Publishing Group UK London}
}

@article{sharma2019deepinsight,
  title={DeepInsight: A methodology to transform a non-image data to an image for convolution neural network architecture},
  author={Sharma, Alok and Vans, Edwin and Shigemizu, Daichi and Boroevich, Keith A and Tsunoda, Tatsuhiko},
  journal={Scientific reports},
  volume={9},
  number={1},
  pages={11399},
  year={2019},
  publisher={Nature Publishing Group UK London}
}

@article{sharma2024enhanced,
  title={Enhanced analysis of tabular data through Multi-representation DeepInsight},
  author={Sharma, Alok and L{\'o}pez, Yosvany and Jia, Shangru and Lysenko, Artem and Boroevich, Keith A and Tsunoda, Tatsuhiko},
  journal={Scientific Reports},
  volume={14},
  number={1},
  pages={12851},
  year={2024},
  publisher={Nature Publishing Group UK London}
}

@article{sharma2023deepinsight,
  title={DeepInsight-3D architecture for anti-cancer drug response prediction with deep-learning on multi-omics},
  author={Sharma, Alok and Lysenko, Artem and Boroevich, Keith A and Tsunoda, Tatsuhiko},
  journal={Scientific reports},
  volume={13},
  number={1},
  pages={2483},
  year={2023},
  publisher={Nature Publishing Group UK London}
}

@article{sharma2022classification,
  title={Classification with 2-D convolutional neural networks for breast cancer diagnosis},
  author={Sharma, Anuraganand and Kumar, Dinesh},
  journal={Scientific Reports},
  volume={12},
  number={1},
  pages={21857},
  year={2022},
  publisher={Nature Publishing Group UK London}
}

@article{borisov2022deep,
  title={Deep neural networks and tabular data: A survey},
  author={Borisov, Vadim and Leemann, Tobias and Se{\ss}ler, Kathrin and Haug, Johannes and Pawelczyk, Martin and Kasneci, Gjergji},
  journal={IEEE transactions on neural networks and learning systems},
  volume={35},
  number={6},
  pages={7499--7519},
  year={2022},
  publisher={IEEE}
}

@article{castillo2023tinto,
  title={TINTO: converting tidy data into image for classification with 2-dimensional convolutional neural networks},
  author={Castillo-Cara, Manuel and Talla-Chumpitaz, Reewos and Garc{\'\i}a-Castro, Ra{\'u}l and Orozco-Barbosa, Luis},
  journal={SoftwareX},
  volume={22},
  pages={101391},
  year={2023},
  publisher={Elsevier}
}

@article{azizi2024review,
  title={A review of visibility graph analysis},
  author={Azizi, Hadis and Sulaimany, Sadegh},
  journal={IEEE Access},
  volume={12},
  pages={93517--93530},
  year={2024},
  publisher={IEEE}
}

@article{bazgir2020representation,
  title={Representation of features as images with neighborhood dependencies for compatibility with convolutional neural networks},
  author={Bazgir, Omid and Zhang, Ruibo and Dhruba, Saugato Rahman and Rahman, Raziur and Ghosh, Souparno and Pal, Ranadip},
  journal={Nature communications},
  volume={11},
  number={1},
  pages={4391},
  year={2020},
  publisher={Nature Publishing Group UK London}
}

@article{talla2023novel,
  title={A novel deep learning approach using blurring image techniques for Bluetooth-based indoor localisation},
  author={Talla-Chumpitaz, Reewos and Castillo-Cara, Manuel and Orozco-Barbosa, Luis and Garc{\'\i}a-Castro, Ra{\'u}l},
  journal={Information Fusion},
  volume={91},
  pages={173--186},
  year={2023},
  publisher={Elsevier}
}

@article{liu2026interpretable,
  title={Interpretable Hybrid Vision Transformer Architectures for MIMO-Based Indoor Localization using Synthetic Spatial Representations},
  author={Liu, Jiayun and Castillo-Cara, Manuel and Garc{\'\i}a-Castro, Ra{\'u}l and Orozco-Barbosa, Luis},
  journal={IEEE Internet of Things Journal},
  year={2026},
  publisher={IEEE}
}

@article{castillo2025mimo,
  title={MIMO-Based Indoor Localisation with Hybrid Neural Networks: Leveraging Synthetic Images from Tidy Data for Enhanced Deep Learning},
  author={Castillo-Cara, Manuel and Mart{\'\i}nez-G{\'o}mez, Jesus and Ballesteros-Jerez, Javier and Garc{\'\i}a-Varea, Ismael and Garc{\'\i}a-Castro, Ra{\'u}l and Orozco-Barbosa, Luis},
  journal={IEEE Journal of Selected Topics in Signal Processing},
  year={2025},
  publisher={IEEE}
}

@article{mamdouh2026tab2visual,
  title={Tab2Visual: Deep Learning for Limited Tabular Data via Visual Representations and Augmentation},
  author={Mamdouh, Ahmed and El-Melegy, Moumen and Ali, Samia and Kikinis, Ron},
  journal={Pattern Recognition},
  pages={113173},
  year={2026},
  publisher={Elsevier}
}

@article{alenizy2025transforming,
  title={Transforming tabular data into images via enhanced spatial relationships for CNN processing},
  author={Alenizy, Hameedah A and Berri, Jawad},
  journal={Scientific Reports},
  volume={15},
  number={1},
  pages={17004},
  year={2025},
  publisher={Nature Publishing Group UK London}
}

@article{damri2024towards,
  title={Towards efficient image-based representation of tabular data},
  author={Damri, Amit and Last, Mark and Cohen, Niv},
  journal={Neural Computing and Applications},
  volume={36},
  number={2},
  pages={1023--1043},
  year={2024},
  publisher={Springer}
}

@article{briner2023tabular,
  title={Tabular-to-image transformations for the classification of anonymous network traffic using deep residual networks},
  author={Briner, Nathan and Cullen, Drake and Halladay, James and Miller, Darrin and Primeau, Riley and Avila, Abraham and Basnet, Ram and Doleck, Tenzin},
  journal={IEEE Access},
  volume={11},
  pages={113100--113113},
  year={2023},
  publisher={IEEE}
}

@article{lara2025transfer,
  title={Transfer learning for a tabular-to-image approach: A case study for cardiovascular disease prediction},
  author={Lara-Abelenda, Francisco J and Chushig-Muzo, David and Peiro-Corbacho, Pablo and G{\'o}mez-Mart{\'\i}nez, Vanesa and W{\"a}gner, Ana M and Granja, Concei{\c{c}}{\~a}o and Soguero-Ruiz, Cristina},
  journal={Journal of Biomedical Informatics},
  volume={165},
  pages={104821},
  year={2025},
  publisher={Elsevier}
}

@inproceedings{selvaraju2017grad,
  title={Grad-cam: Visual explanations from deep networks via gradient-based localization},
  author={Selvaraju, Ramprasaath R and Cogswell, Michael and Das, Abhishek and Vedantam, Ramakrishna and Parikh, Devi and Batra, Dhruv},
  booktitle={Proceedings of the IEEE international conference on computer vision},
  pages={618--626},
  year={2017}
}

@article{medeiros2023comparative,
  title={A comparative analysis of converters of tabular data into image for the classification of Arboviruses using Convolutional Neural Networks},
  author={Medeiros Neto, Leonides and Rogerio da Silva Neto, Sebasti{\~a}o and Endo, Patricia Takako},
  journal={Plos one},
  volume={18},
  number={12},
  pages={e0295598},
  year={2023},
  publisher={Public Library of Science San Francisco, CA USA}
}

@article{lee2024table2image,
  title={Table2Image: interpretable tabular data classification with realistic image transformations},
  author={Lee, Seungeun and Kwak, Il-Youp and Lee, Kihwan and Bae, Subin and Lee, Sangjun and Lee, Seulbin and Oh, Seungsang},
  journal={arXiv preprint arXiv:2412.06265},
  year={2024}
}



%








\end{document}